# Star-Graph Multimodal Matching Component Analysis for Data Fusion and Transfer Learning

Nick Lorenzo*

August 5, 2022

**Abstract**

Previous matching component analysis (MCA) techniques map two data domains to a common domain for further processing in data fusion and transfer learning contexts. In this paper, we extend these techniques to the star-graph multimodal (SGM) case in which one particular data domain is connected to $m$ others via an objective function. We provide a particular feasible point for the resulting trace maximization problem in closed form and algorithms for its computation and iterative improvement, leading to our main result, the SGM maps. We also provide numerical examples demonstrating that SGM is capable of encoding into its maps more information than MCA when few training points are available. In addition, we develop a further generalization of the MCA covariance constraint, eliminating a previous feasibility condition and allowing larger values of the rank of the prescribed covariance matrix.

## 1 Introduction

The matching component analysis (MCA) technique for transfer learning [1] finds two maps – one from each of two data domains to a lower-dimensional, common domain – using only a small number of matched data pairs, where each matched data pair is comprised of one point from each data domain. These maps minimize the expected distance between mapped data pairs within the common domain, subject to an identity matrix covariance constraint and an affine linear structure. Learning techniques can then be applied to matched data points after they are mapped to the common domain, where each such point is encoded with information from both data domains via its respective optimal affine linear transformation.

In [2], the covariance-generalized MCA (CGMCA) technique was developed in order to allow for the encoding of additional statistical information into the MCA maps. This was done by generalizing the identity matrix covariance constraint of MCA to accommodate any covariance matrix (compare Figures 1a and 1b).

We are interested in extending the application space of CGMCA to accommodate three or more data domains simultaneously. In this paper, we restrict our attention to the star-graph case of this multimodal[1] generalization: one central modality is mapped into the common domain as closely as possible to each of the other modalities, with no two non-central modalities mapped together in this way. These close-mapping conditions appear as norms in the objective function (5a), and this norm-connectedness is represented schematically by dotted lines in Figure 1. We refer to this star-graph multimodal generalization of CGMCA as SGM.

Multimodal data can already be handled one way by MCA and CGMCA: matched, vectorized data points from two or more modalities can be stacked on top of one another, forming higher-dimensional data points [3]. These higher-dimensional data points can then be treated as though they come from a single modality, with MCA or CGMCA applied as usual; this is illustrated in Figure 4c, where the first modality is unstacked and two sub-modalities are stacked into the second modality. This approach, however, provides no immediately apparent mechanism for weighting the influence of each sub-modality in the resulting maps. It also effectively forces each set of stacked sub-modalities to directly compete with one another for representation in the common domain, as all stacked sub-modalities are represented by a single MCA map. In addition, this approach enforces a single covariance structure onto the common-domain image of the stacked sub-modalities, rather than allowing each sub-modality to have its own prescribed covariance. In order to overcome these limitations, we develop SGM.

---

*University of Dayton Research Institute, Dayton, OH, USA (`Nicholas.Lorenzo@udri.udayton.edu`)
[1] We use the terms "data domain" and "modality" interchangeably.

## 2 Mathematical development of SGM

Our development of the SGM optimization problem mirrors the development of the MCA [1] and CGMCA [2] optimization problems, with some notation borrowed from [4, p. 130].

### 2.1 Notation

For $x \in \mathbb{N}$, let $[x] \equiv \{1, \ldots, x\}$ and let $[x]_0 \equiv \{0, \ldots, x\} = [x] \cup \{0\}$.

Let $a, b, c \in \mathbb{N}$ with $c \leq \min\{a, b\}$, $v \in \mathbb{R}^c$, and $A \in \mathbb{R}^{a \times b}$. Define

$$\operatorname{diag}_{a \times b}(v) \in \mathbb{R}^{a \times b} \tag{1a}$$

to be the matrix whose first $c$ diagonal elements are the elements of $v$ (in order) and zero elsewhere, and define

$$\operatorname{diag}_c^{-1}(A) \in \mathbb{R}^c \tag{1b}$$

to be the vector comprised of the first $c$ diagonal elements of the matrix $A$ (in order). We also define

$$\operatorname{dg}_c(A) \equiv \operatorname{diag}_{c \times c}\left(\operatorname{diag}_c^{-1}(A)\right) \in \mathbb{R}^{c \times c} \tag{1c}$$

to be the diagonal matrix whose diagonal entries are the first $c$ diagonal elements of the matrix $A$.

Let

$$\mathbb{O}^{a \times b} \equiv \{O \in \mathbb{R}^{a \times b} \mid O^\mathsf{T} O = I_b\} \tag{2a}$$

denote the set of $a \times b$ real matrices with orthonormal columns (this set is non-empty only if $b \leq a$; when $b = a$, these matrices are orthogonal), let

$$\mathbb{D}_+^a \equiv \{D \in \mathbb{R}^{a \times a} \mid D = \operatorname{diag}_{a \times a}\left([\delta_1, \ldots, \delta_a]^\mathsf{T}\right) \; \exists \; \delta_1 \geq \ldots \geq \delta_a > 0\} \tag{2b}$$

denote the set of diagonal $a \times a$ real invertible matrices with non-increasing elements, and let

$$\mathbb{D}_{\not{}}^{a \times b} \equiv \{D \in \mathbb{R}^{a \times b} \mid D = \operatorname{diag}_{a \times b}\left([\delta_1, \ldots, \delta_{\min\{a,b\}}]^\mathsf{T}\right) \; \exists \; \delta_1 \geq \ldots \geq \delta_{\min\{a,b\}} \geq 0\} \tag{2c}$$

denote the set of diagonal $a \times b$ real matrices with non-negative, non-increasing elements.

We reserve the undecorated but subscripted symbols $U_{(\cdot)}$, $\Sigma_{(\cdot)}$, and $V_{(\cdot)}$ and barred, subscripted symbols $\bar{U}_{(\cdot)}$, $\bar{\Sigma}_{(\cdot)}$, and $\bar{V}_{(\cdot)}$ exclusively for use as factors in the SVD. In particular, for a matrix $Z \in \mathbb{R}^{\operatorname{rows}(Z) \times \operatorname{cols}(Z)}$, we denote an arbitrary but fixed thin SVD of $Z$ by

$$Z = U_Z \Sigma_Z V_Z^\mathsf{T}, \qquad \text{where} \qquad U_Z \in \mathbb{O}^{\operatorname{rows}(Z) \times \operatorname{rank}(Z)}, \quad \Sigma_Z \in \mathbb{D}_+^{\operatorname{rank}(Z)}, \quad V_Z \in \mathbb{O}^{\operatorname{cols}(Z) \times \operatorname{rank}(Z)}, \tag{3a}$$

and we denote a fixed full SVD of $Z$ by

$$Z = \bar{U}_Z \bar{\Sigma}_Z \bar{V}_Z^\mathsf{T}, \qquad \text{where} \qquad \bar{U}_Z \in \mathbb{O}^{\operatorname{rows}(Z) \times \operatorname{rows}(Z)}, \quad \bar{\Sigma}_Z \in \mathbb{D}_{\not{}}^{\operatorname{rows}(Z) \times \operatorname{cols}(Z)}, \quad \bar{V}_Z \in \mathbb{O}^{\operatorname{cols}(Z) \times \operatorname{cols}(Z)}, \tag{3b}$$

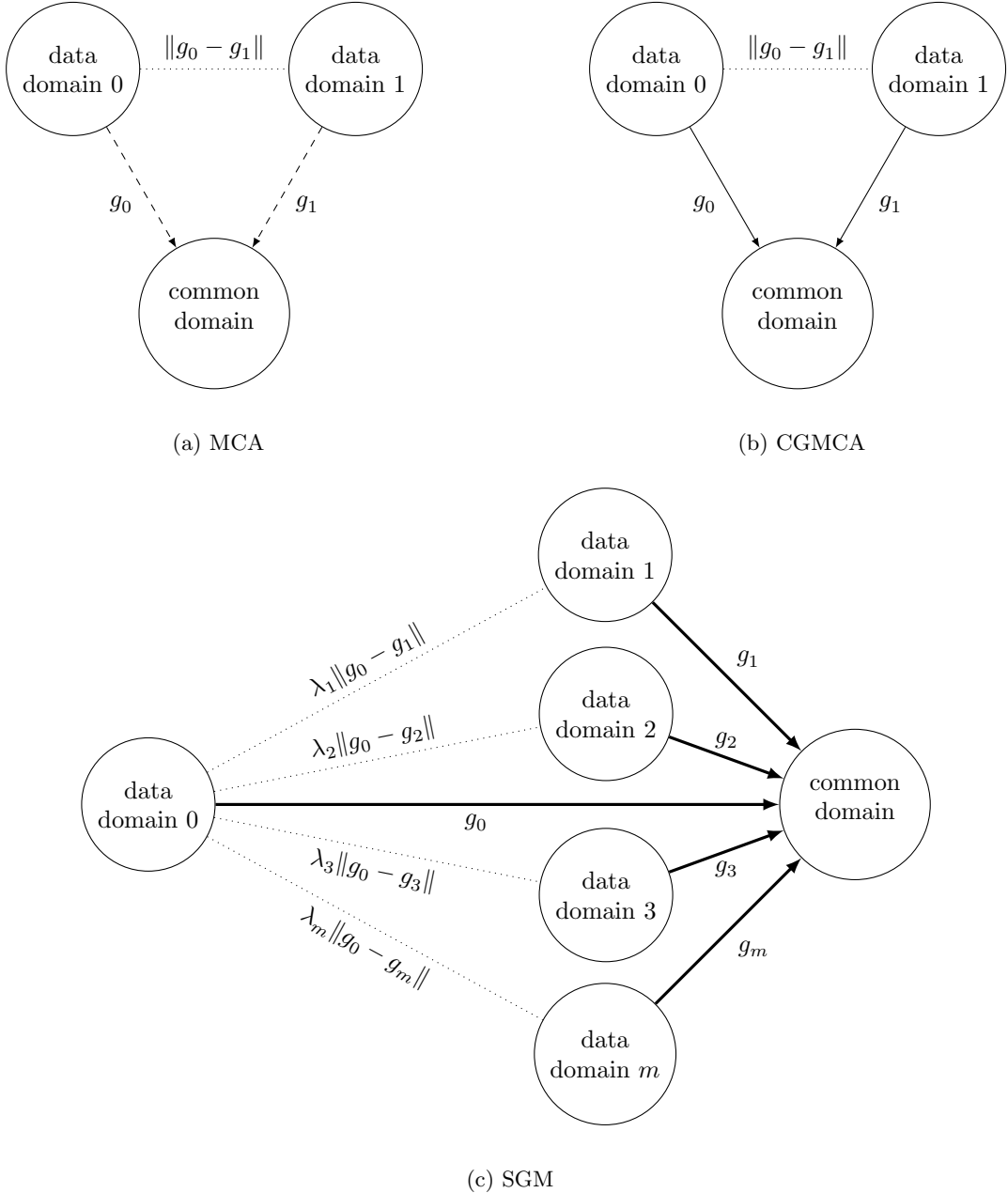

Figure 1: Schematic representation of the norm-connectedness and mappings of (a) MCA, (b) CGMCA, and (c) SGM. Each data domain maps into the common domain (arrows); data domain 0 is norm-connected to each of the other data domains (dotted lines). The dashed arrows for MCA indicate its restriction to an identity covariance constraint; the solid arrows for CGMCA indicate its ability to prescribe covariance information; the thick arrows for SGM represent its further generalized covariance constraint (see Section 2.4). In SGM, the dotted lines form a star graph with data domain 0 at its center.

with $\bar{U}_Z$ an arbitrary but fixed orthogonal completion of $U_Z$, with $\bar{V}_Z$ an arbitrary but fixed orthogonal completion of $V_Z$, and with $\bar{\Sigma}_Z$ having $\Sigma_Z$ as a leading principal submatrix with zeros elsewhere.

We also reserve the subscripted and superscripted symbol $B_{(\cdot)}^{(\cdot)}$ to denote by $B_Z^\ell$ the best approximation (in the Frobenius norm) of $Z$ having rank at most $\ell$ [5, p. 36], where $\ell \in [\text{rank}(Z)]$:

$$B_Z^\ell \equiv \sum_{j \in [\ell]} (\Sigma_Z)_{jj} (U_Z)_j (V_Z)_j^\mathsf{T} \in \mathbb{R}^{\text{rows}(Z) \times \text{cols}(Z)}, \tag{4a}$$

with $(\,\cdot\,)_j$ denoting the $j^{\text{th}}$ column of a matrix and where

$$U_{B_Z^\ell} = (U_Z)_{[\ell]} \in \mathbb{O}^{\text{rows}(Z) \times \ell}, \quad \Sigma_{B_Z^\ell} = \text{dg}_\ell(\Sigma_Z) \in \mathbb{D}_+^\ell, \quad V_{B_Z^\ell} = (V_Z)_{[\ell]} \in \mathbb{O}^{\text{cols}(Z) \times \ell}, \tag{4b}$$

with $(\,\cdot\,)_{[\ell]}$ denoting the first $\ell$ columns of a matrix.

## 2.2 Problem statement

For some $m \in \mathbb{N}$, we consider random variables $\{X_i\}_{i \in [m]_0}$, each with domain probability space $(\Omega, \mathcal{M}, P)$, and a set of outcomes $\{\omega_j\}_{j \in [n]} \subseteq \Omega$ for some $n \in \mathbb{N}$ with $1 < n$, and we are provided, for each $i \in [m]_0$, with the data set $\{X_i(\omega_j)\}_{j \in [n]}$ of independent random variates belonging to the $i^{\text{th}}$ data domain $\Omega_{X_i} = \mathbb{R}^{d_i}$ for some $d_i \in \mathbb{N}$ with $1 < d_i$. For each $j \in [n]$, we refer to the length-$(m+1)$ tuple of realizations $(X_0(\omega_j), \ldots, X_m(\omega_j))$ as a matched data point. Provided a covariance matrix $C_i C_i^\mathsf{T} \in \mathbb{R}^{k \times k}$ for each $i \in [m]_0$ and for some fixed $k \in \mathbb{N}$, we wish to use the matched data to approximately solve the weighted, constrained optimization problem

$$\text{minimize} \quad \mathbb{E}\left[\sum_{i \in [m]} \lambda_i \|g_0(X_0) - g_i(X_i)\|_2^2\right], \tag{5a}$$

$$\text{subject to} \quad g_i : \mathbb{R}^{d_i} \to \mathbb{R}^k, \tag{5b}$$

$$\mathbb{E}\, g_i(X_i) = 0, \tag{5c}$$

$$\mathbb{E}\left[g_i(X_i) g_i(X_i)^\mathsf{T}\right] = C_i C_i^\mathsf{T}, \tag{5d}$$

$$i \in [m]_0,$$

where the weights $\{\lambda_i\}_{i \in [m]}$ are fixed, non-negative, and sum to unity. The fact that the map $g_0$ is paired with each of the other maps $\{g_i\}_{i \in [m]}$ in the objective function (5a) is reflected by the dotted lines of Figure 1c.

The optimization problem (5) is our multimodal generalization of the optimization problem (1) of [2] (in which $m \equiv 1$, represented schematically in Figure 1b), itself a covariance-generalized version of the optimization problem (2.1) of [1] (in which both $m \equiv 1$ and $C_i C_i^\mathsf{T} \equiv I_k$, represented schematically in Figure 1a); we seek to minimize the sum of weighted expected distances between the images under all pairs $\{(g_0, g_i)\}_{i \in [m]}$ of the random variables $\{(X_0, X_i)\}_{i \in [m]}$ within the common domain of dimension $k$, while centering the resulting mappings $\{g_i\}_{i \in [m]_0}$ via (5c) and preserving the prescribed covariance structures $\{C_i C_i^\mathsf{T}\}_{i \in [m]_0}$ of (5d).

In this star-graph version of the problem, we have one modality (indexed as the $0^{\text{th}}$ modality) at the center of the star graph, with $m$ other modalities each norm-connected to the center modality but not to each other. Thus the map $g_0$ is directly influenced by each of the maps $\{g_i\}_{i \in [m]}$, while the maps $\{g_i\}_{i \in [m]}$ have no direct influence over one another (see the dotted lines of Figure 1c for a schematic representation of this norm-connectedness).

## 2.3 Approximation via available data

Restricting $g_i$ to be an affine linear transformation, the requirement (5b) is equivalent to the requirement that $g_i(x) = A_i x + b_i$ for some $A_i \in \mathbb{R}^{k \times d_i}$ and $b_i \in \mathbb{R}^k$. Denoting by $\hat{\mu}_i$ and $\hat{\Theta}_i$ the mean and covariance, respectively, of $X_i$, the centering constraint (5c) then implies the condition $b_i = -A_i \hat{\mu}_i$, so that

$$g_i(x) = A_i(x - \hat{\mu}_i). \tag{6}$$

The covariance constraint (5d) then becomes the condition $A_i \hat{\Theta}_i A_i^\mathsf{T} = C_i C_i^\mathsf{T}$, and the optimization problem (5) becomes the optimization problem

$$\text{minimize} \quad \mathbb{E}\left[\sum_{i \in [m]} \lambda_i \|A_0(X_0 - \hat{\mu}_0) - A_i(X_i - \hat{\mu}_i)\|_2^2\right], \tag{7a}$$

$$\text{subject to} \quad A_i \in \mathbb{R}^{k \times d_i}, \tag{7b}$$

$$A_i \hat{\Theta}_i A_i^\mathsf{T} = C_i C_i^\mathsf{T}, \tag{7c}$$

$$i \in [m]_0.$$

Using the $n$ realizations of each $X_i$ to which we have access, we approximate the true means $\hat{\mu}_i$ and true covariances $\hat{\Theta}_i$ via the sample means $\mu_i$ and sample covariances $\Theta_i$:

$$\hat{\mu}_i \approx \mu_i \equiv \frac{1}{n} \sum_{j \in [n]} X_i(\omega_j) \in \mathbb{R}^{d_i}, \tag{8a}$$

$$\hat{\Theta}_i \approx \Theta_i \equiv S_i S_i^\mathsf{T} \in \mathbb{R}^{d_i \times d_i}, \tag{8b}$$

where we define $S_i \in \mathbb{R}^{d_i \times n}$ to be the matrix whose $j^{\text{th}}$ column is the vector

$$(S_i)_j \equiv \frac{1}{\sqrt{n-1}}(X_i(\omega_j) - \mu_i) \in \mathbb{R}^{d_i}. \tag{9}$$

Then the optimization problem (7) is approximated by the optimization problem

$$\text{minimize} \quad \frac{n-1}{n} \sum_{j \in [n]} \sum_{i \in [m]} \lambda_i \left\| A_0(S_0)_j - A_i(S_i)_j \right\|_2^2, \tag{10a}$$

$$\text{subject to} \quad A_i \in \mathbb{R}^{k \times d_i}, \tag{10b}$$

$$A_i \Theta_i A_i^\mathsf{T} = C_i C_i^\mathsf{T}, \tag{10c}$$

$$i \in [m]_0,$$

which is equivalent to the optimization problem

$$\text{minimize} \quad \frac{n-1}{n} \sum_{i \in [m]} \lambda_i \left\| A_0 S_0 - A_i S_i \right\|_F^2, \tag{11a}$$

$$\text{subject to} \quad A_i \in \mathbb{R}^{k \times d_i}, \tag{11b}$$

$$A_i \Theta_i A_i^\mathsf{T} = C_i C_i^\mathsf{T}, \tag{11c}$$

$$i \in [m]_0.$$

## 2.4 A further generalization of the covariance constraint

The covariance constraint (11c) implies infeasibility unless

$$\text{rank}(S_i) \geq \text{rank}(C_i C_i^\mathsf{T}); \tag{12}$$

since $\Theta_i$ is a sample covariance matrix, we also find that

$$\text{rank}(\Theta_i) = \text{rank}(S_i) \leq \min\{d_i, n-1\}. \tag{13}$$

In addition, we note that

$$\text{rank}(C_i C_i^\mathsf{T}) \leq k. \tag{14}$$

In order to allow for cases where the feasibility requirement (12) is not satisfied, we consider the further generalized covariance constraint

$$A_i \in \arg \min_{\substack{\hat{A}_i \in \mathbb{R}^{k \times d_i} \\ \operatorname{rank}(\hat{A}_i) \leq \operatorname{rank}(S_i)}} \|\hat{A}_i \Theta_i \hat{A}_i^\mathsf{T} - C_i C_i^\mathsf{T}\|_F, \tag{15}$$

which requires $A_i \Theta_i A_i^\mathsf{T}$ to be as close as possible to the prescribed covariance matrix $C_i C_i^\mathsf{T}$, even when equality is not possible. Here, the rank restriction $\operatorname{rank}(\hat{A}_i) \leq \operatorname{rank}(S_i)$ was chosen because any product of matrices with $\Theta_i$ as a factor has rank at most $\operatorname{rank}(S_i)$. In cases where $\operatorname{rank}(S_i) < \operatorname{rank}(C_i C_i^\mathsf{T})$, so that the feasibility requirement (12) is not satisfied, the constraint (15) effectively replaces $C_i C_i^\mathsf{T}$ with its best approximation of rank $\operatorname{rank}(S_i)$. With this in mind, we may write the constraint (15) simply as

$$A_i \Theta_i A_i^\mathsf{T} = B_{C_i C_i^\mathsf{T}}^{r_i^{\min}}, \tag{16}$$

where we have used the notation (4a). Here,

$$r_i^{\min} \equiv \min\{\operatorname{rank}(C_i C_i^\mathsf{T}), \operatorname{rank}(S_i)\}. \tag{17}$$

With the constraint (16) replacing the constraint (11c), the optimization problem (11) becomes the optimization problem

$$\text{minimize} \quad \frac{n-1}{n} \sum_{i \in [m]} \lambda_i \|A_0 S_0 - A_i S_i\|_F^2, \tag{18a}$$

$$\text{subject to} \quad A_i \in \mathbb{R}^{k \times d_i}, \tag{18b}$$

$$A_i \Theta_i A_i^\mathsf{T} = B_{C_i C_i^\mathsf{T}}^{r_i^{\min}}, \tag{18c}$$

$$i \in [m]_0,$$

where the feasibility requirement (12) no longer applies.

## 2.5 Transformation to trace maximization problem

In view of (4b), we find that

$$B_{C_i C_i^\mathsf{T}}^{r_i^{\min}} = (U_{C_i})_{[r_i^{\min}]} \operatorname{dg}_{r_i^{\min}}(\Sigma_{C_i}^2) \left((U_{C_i})_{[r_i^{\min}]}\right)^\mathsf{T}, \tag{19}$$

where we have used the notation (1c). Using the result (19), we make the change of variables

$$A_i = (U_{C_i})_{[r_i^{\min}]} \operatorname{dg}_{r_i^{\min}}(\Sigma_{C_i}) D_i \Sigma_{S_i}^{-1} U_{S_i}^\mathsf{T}, \tag{20a}$$

where the semi-orthogonal matrix

$$D_i^\mathsf{T} \in \mathbb{O}^{\operatorname{rank}(S_i) \times r_i^{\min}} \tag{20b}$$

is unknown. With $A_i$ defined as in (20a), the covariance constraint (18c) is satisfied for each $D_i$ satisfying (20b).

Using the change of variables (20a), we find that

$$A_i S_i = (U_{C_i})_{[r_i^{\min}]} \operatorname{dg}_{r_i^{\min}}(\Sigma_{C_i}) D_i V_{S_i}^\mathsf{T} \tag{21}$$

and

$$\operatorname{tr}\left((A_j S_j)(A_i S_i)^\mathsf{T}\right) = \operatorname{tr}\left((U_{C_j})_{[r_j^{\min}]} \operatorname{dg}_{r_j^{\min}}(\Sigma_{C_j}) D_j V_{S_j}^\mathsf{T} V_{S_i} D_i^\mathsf{T} \operatorname{dg}_{r_i^{\min}}(\Sigma_{C_i}) ((U_{C_i})_{[r_i^{\min}]})^\mathsf{T}\right) \tag{22a}$$

$$= \operatorname{tr}\left(D_i^\mathsf{T} \operatorname{dg}_{r_i^{\min}}(\Sigma_{C_i})((U_{C_i})_{[r_i^{\min}]})^\mathsf{T}(U_{C_j})_{[r_j^{\min}]}\operatorname{dg}_{r_j^{\min}}(\Sigma_{C_j})D_j V_{S_j}^\mathsf{T} V_{S_i}\right). \tag{22b}$$

Then each norm in the objective function (18a) has the form

$$\|A_0 S_0 - A_i S_i\|_F^2 = \operatorname{tr}\left((A_0 S_0)(A_0 S_0)^\mathsf{T}\right) + \operatorname{tr}\left((A_i S_i)(A_i S_i)^\mathsf{T}\right) - 2\operatorname{tr}\left((A_0 S_0)(A_i S_i)^\mathsf{T}\right) \tag{23a}$$

$$= \operatorname{tr}\left(B_{C_0 C_0^\mathsf{T}}^{r_0^{\min}} + B_{C_i C_i^\mathsf{T}}^{r_i^{\min}}\right) \tag{23b}$$
$$- 2\operatorname{tr}\left(D_i^\mathsf{T} \operatorname{dg}_{r_i^{\min}}(\Sigma_{C_i})((U_{C_i})_{[r_i^{\min}]})^\mathsf{T}(U_{C_0})_{[r_0^{\min}]}\operatorname{dg}_{r_0^{\min}}(\Sigma_{C_0})D_0 V_{S_0}^\mathsf{T} V_{S_i}\right).$$

The optimization problem (18) can then be solved by finding

$$\mathcal{D}^{\mathrm{opt}} \in \arg\max_{\mathcal{D}\in\mathcal{F}} \sum_{i\in[m]} \lambda_i \operatorname{tr}\left(D_i^\mathsf{T} \operatorname{dg}_{r_i^{\min}}(\Sigma_{C_i})((U_{C_i})_{[r_i^{\min}]})^\mathsf{T}(U_{C_0})_{[r_0^{\min}]}\operatorname{dg}_{r_0^{\min}}(\Sigma_{C_0})D_0 V_{S_0}^\mathsf{T} V_{S_i}\right), \tag{24a}$$

where

$$\mathcal{F} \equiv \{(D_0,\ldots,D_m) \mid D_i \text{ satisfies (20b)}, i \in [m]_0\} \tag{24b}$$

is the set of length-$(m+1)$ tuples of feasible matrices.

In its exact form, this problem can be cast as a non-convex quadratically constrained quadratic program (QCQP) over a product Stiefel manifold. In general, this class of problems has no known exact solution [7]. Due to the nature of the product manifold involved, we note that the pursuit of a purely numerical solution is a difficult problem, even using specialized manifold optimization packages such as [10]. Rather than pursue a purely numerical solution, then, we offer an approximate solution as follows. In Appendix A, we obtain an exact solution to a related problem; this exact solution is feasible for the original problem. In Appendix B, we provide an iterative algorithm to potentially improve upon any feasible point while maintaining feasibility. In Section 2.6, we combine these processes to obtain a particular feasible point for the optimization problem of interest.

## 2.6 A particular feasible point for the trace maximization problem

Applying the particular solutions (36) and (38) of Appendix A to the objective function (24a), we find that

$$D_0^{\mathrm{p}} = \bar{V}_H \left((\bar{U}_H)_{[r_0^{\min}]}\right)^\mathsf{T}, \tag{25a}$$

$$D_i^{\mathrm{p}} = \bar{V}_{L_i} \left((\bar{V}_{R_i})_{[r_i^{\min}]}\right)^\mathsf{T}, \quad i \in [m], \tag{25b}$$

with

$$L_i \equiv \operatorname{dg}_{r_0^{\min}}(\Sigma_{C_0})((U_{C_0})_{[r_0^{\min}]})^\mathsf{T}(U_{C_i})_{[r_i^{\min}]}\operatorname{dg}_{r_i^{\min}}(\Sigma_{C_i}), \quad R_i \equiv V_{S_0}^\mathsf{T} V_{S_i}, \tag{25c}$$

and

$$H \equiv \sum_{i\in[m]} \lambda_i R_i (D_i^{\mathrm{p}})^\mathsf{T} L_i^\mathsf{T}. \tag{25d}$$

We may then apply Algorithm B.1 of Appendix B to the particular solutions (25) in order to obtain the particular feasible point

$$(D_0^{\mathrm{iter}},\ldots,D_m^{\mathrm{iter}}) \equiv \mathcal{A}_{B.1}((D_0^{\mathrm{p}},\ldots,D_m^{\mathrm{p}}),\{L_i,R_i\}_{i\in[m]},\{\lambda_i\}_{i\in[m]}), \tag{26}$$

where we have used the notation (41).

## 2.7 Reduction of our particular feasible point to an optimal solution in the case of one non-central modality

In the case of one non-central modality, we have $m = 1$. If, in addition, the original feasibility requirement (12) is satisfied, then the SGM optimization problem (18) reduces to the optimization problem (7) of CGMCA [2]. In this case, $L_1$ of (25c) reduces to $A$ of (34b) of [2], $D_1^{\text{P}}$ of (25b) reduces to $D_2^{\text{opt}}$ of (34a) of [2], $H$ of (25d) reduces to $\bar{U}_{R_1}\bar{\Sigma}_{R_1}\bar{V}_{R_1}^{\mathsf{T}}(\bar{V}_{R_1})_{[r_1^{\text{cov}}]}\bar{\Sigma}_{L_1}^{\mathsf{T}}\bar{U}_{L_1}^{\mathsf{T}}$, and $D_0^{\text{P}}$ of (25a) reduces to $\bar{V}_{L_1}((\bar{U}_{R_1})_{[r_0^{\text{cov}}]})^{\mathsf{T}}$, which matches (34a) of [2]. Thus $D_0^{\text{P}}$ and $D_1^{\text{P}}$ match the CGMCA solutions (34a) of [2], and Algorithm B.1 will leave these solutions unchanged, as they are already optimal. In this way, SGM is a genuine generalization of CGMCA.

## 3 Main results

Here we state our main results.

### 3.1 SGM maps

Substituting the particular feasible point (26) into the change of variables (20a) and using the approximation (8a) in the result (6), we find particular affine linear transformations for the original optimization problem (5) to be given by

$$g_i^{\text{SGM}}(x) = A_i^{\text{SGM}} x + b_i^{\text{SGM}}, \tag{27a}$$

where

$$A_i^{\text{SGM}} \equiv (U_{C_i})_{[r_i^{\min}]} \operatorname{dg}_{r_i^{\min}}(\Sigma_{C_i}) D_i^{\text{iter}} \Sigma_{S_i}^{-1} U_{S_i}^{\mathsf{T}}, \quad b_i^{\text{SGM}} \equiv -A_i^{\text{SGM}} \mu_i, \quad i \in [m]_0. \tag{27b}$$

### 3.2 SGM algorithm

A method to compute the SGMCGMCA feasible point (27) from the data, prescribed covariance matrices, and weights is provided in Algorithm 3.1 below.

Algorithm 3.1: Star-Graph Multimodal MCA.

---

STAR-GRAPH MULTIMODAL MCA

---

1 **Inputs:** data $\{\{X_i(\omega_j)\}_{j\in[n]}\}_{i\in[m]_0}$, covariances $\{C_i C_i^\mathsf{T} \in \mathbb{R}^{k\times k}\}_{i\in[m]_0}$, and weights $\{\lambda_i\}_{i\in[m]}$

2 **for** $i \in [m]_0$

3     Compute sample mean $\mu_i$      See (8a)

4     Compute scaled and centered data matrix $S_i$      See (9)

5     Compute thin SVDs $S_i = U_{S_i} \Sigma_{S_i} V_{S_i}^\mathsf{T}$ and $C_i C_i^\mathsf{T} = U_{C_i} \Sigma_{C_i}^2 U_{C_i}^\mathsf{T}$

6     Compute matrix $\Sigma_{C_i} = (\Sigma_{C_i}^2)^{1/2}$

7 **end**

8 **for** $i \in [m]$

9     Compute matrices $L_i$ and $R_i$      See (25c)

10 **end**

11 $(D_0^{\mathrm{p}}, \ldots, D_m^{\mathrm{p}}) \leftarrow \mathcal{A}_{A.1}(\{L_i, R_i\}_{i\in[m]}, \{\lambda_i\}_{i\in[m]})$      See (39)

12 $(D_0^{\mathrm{iter}}, \ldots, D_m^{\mathrm{iter}}) \leftarrow \mathcal{A}_{B.1}((D_0^{\mathrm{p}}, \ldots, D_m^{\mathrm{p}}), \{L_i, R_i\}_{i\in[m]}, \{\lambda_i\}_{i\in[m]})$      See (41)

13 **for** $i \in [m]_0$

14     Compute linear map $A_i^{\mathrm{SGM}} = (U_{C_i})_{[r_i^{\min}]} \, \mathrm{dg}_{r_i^{\min}}(\Sigma_{C_i}) D_i^{\mathrm{iter}} \Sigma_{S_i}^{-1} U_{S_i}^\mathsf{T}$      See (27)

15     Compute translation $b_i^{\mathrm{SGM}} = -A_i^{\mathrm{SGM}} \mu_i$      See (27)

16 **end**

---

## 4 Numerical experiments

In these numerical experiments, we provide evidence that SGM is capable of encoding into its maps more information than MCA, especially when the maps are trained using only a small number of matched data points. We demonstrate this with a transfer learning task performed on the MNIST data set [8].

### 4.1 Data domains

We define three data domains, each consisting of a tile from the top half of each MNIST image. Two of the tiles are the upper-left and upper-right quadrant of each image, while the other tile is the upper-central "quadrant" containing half of each of the other two tiles. See Figure 2 for an illustration of these tiles and their labels; see Figure 3 for an example of each modality tile from a particular MNIST image. We take the labeling of each tile – 0, 1, or 2 – to be its index as a modality, and we take modality 0 to be the central modality of SGM; it is this central modality that is involved in all our performance comparisons. We write the data domains as

$$\mathcal{X}_i \equiv \{X_i(\omega_j)\}_{j\in[N]}, \tag{28}$$

where $i \in \{0, 1, 2\}$ indexes the modality label. Here, $N$ is the total number of tiles for each modality (found from a subset of 10,000 MNIST images, in our experiments) and $X_i(\omega_j) \in \mathbb{R}^{49}$ is the vectorized version of the $(7 \times 7)$-pixel "quadrant" (corresponding to modality $i$) of the $j^{\mathrm{th}}$ MNIST image. Thus a complete matched data point is a triplet of the form $(X_1(\omega_j), X_2(\omega_j), X_3(\omega_j))$ for some $j \in [N]$, with each slot of the triplet corresponding to one vectorized "quadrant" of the $j^{\mathrm{th}}$ MNIST image.

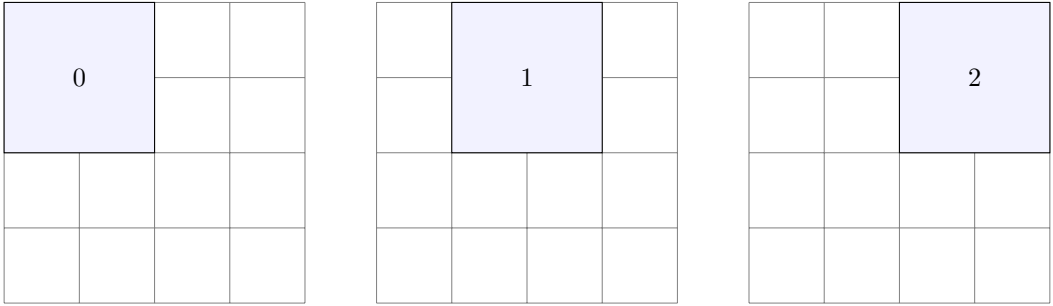

Figure 2: The three tiles (modalities) and corresponding labels used in our experiments. The data set for the $i^{\text{th}}$ modality is written as in (28).

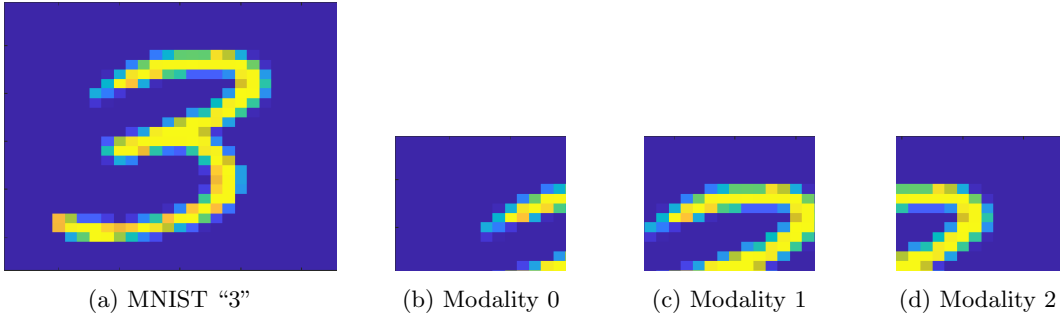

(a) MNIST "3"  (b) Modality 0  (c) Modality 1  (d) Modality 2

Figure 3: An example MNIST image of the digit "3", along with its representation in each of the three modalities of Section 4.1. These modalities match those of Figure 2.

## 4.2 MCA and SGM map computations

Define the sets
$$\mathcal{X}_i^{\text{matched}} \equiv \{X_i(\omega_j)\}_{j \in [n]}, \quad i \in \{0, 1, 2\},  \tag{29}$$
to be the matched data points participating in the MCA and SGM training processes. For each $i$, the elements of $\mathcal{X}_i^{\text{matched}}$ are placed into columns of a data matrix; these data matrices are the inputs to Algorithm 3.1. In order to isolate the effect of the handling of multiple modalities simultaneously, each covariance matrix in each experiment was chosen to be the identity matrix.

In Section 4.3, we summarize the various types of MCA processes used as comparators to SGM's performance; in each case, a data matrix is formed from the matched data points and used as an input to the MCA algorithm [1]. In the case of stacked MCA, the data matrices formed from $\mathcal{X}_1^{\text{matched}}$ and $\mathcal{X}_2^{\text{matched}}$ are vertically stacked into a single data matrix. This training procedure is illustrated at the top of Figure 5.

## 4.3 Testing procedure

After the MCA and SGM maps are trained as in Section 4.2, we compare the performance of these MCA-type techniques to each other and to two baselines.

### 4.3.1 Testing procedure for MCA-type techniques

We test the MCA-type techniques on the following augmented transfer learning task. Some number $n_{\text{kNN}}^{\text{train}}$ of data points from the central domain, including the $n$ used in the formation of the MCA maps, are mapped via $g_0$ to the common domain. A kNN with 15 neighbors is then trained on those central-domain data points. This kNN training procedure is illustrated on the bottom-left of Figure 5.

After the kNN is trained, a number $n_{\text{kNN}}^{\text{test}}$ (chosen to be $2,000$ in all of our experiments) of test data points from the non-central domains are mapped to the common domain, and their weighted common-domain centroid is computed. (Importantly, these test data points correspond to data points not used for any MCA or kNN training.) This kNN testing procedure is illustrated on the bottom-right of Figure 5.

Specifically, define the sets

$$\mathcal{X}_{\text{kNN}}^{\text{train}} \equiv \{X_0(\omega_j)\}_{j \in [n_{\text{kNN}}^{\text{train}}]}, \quad \mathcal{X}_i^{\text{kNN, test}} \equiv \{X_i(\omega_j)\}_{j \in n_{\text{kNN}}^{\text{train}} + [n_{\text{kNN}}^{\text{test}}]}, \quad i \in \{0, 1, 2\}. \tag{30}$$

Here, $n_{\text{kNN}}^{\text{train}} > n$ is the total number of points used in kNN training, and $\mathcal{X}_0^{\text{matched}} \subset \mathcal{X}_{\text{kNN}}^{\text{train}}$. (In all of our experiments, $n_{\text{kNN}}^{\text{train}}$ was chosen to be $5,000$.) For each MCA-type technique, a kNN is trained on the set

$$\{g_0(X_0(\omega_j)) \mid X_0(\omega_j) \in \mathcal{X}_{\text{kNN}}^{\text{train}}\}, \tag{31}$$

the common-domain image of $\mathcal{X}_{\text{kNN}}^{\text{train}}$ under the map $g_0$. Each kNN is then tested on the set

$$\left\{ \sum_{i \in [m]} \lambda_i g_i(X_i(\omega_j)) \mid X_i(\omega_j) \in \mathcal{X}_i^{\text{kNN, test}} \right\}, \tag{32}$$

the weighted centroids of the common-domain images of the test sets $\mathcal{X}_i^{\text{kNN, test}}$ under their respective maps $g_i$ for $i \in [m]$. When $m = 1$ (as is the case for non-SGM techniques), the weighted centroids are just the images under $g_1$ of $\mathcal{X}_1^{\text{kNN, test}}$. Thus each MCA-type technique is trained on domain 0 and tested on the non-central domain(s). This kNN train-test procedure is illustrated on the bottom of Figure 5.

In Section 4.3.3, we discuss exactly which MCA-type techniques are tested in this way.

### 4.3.2 Testing procedure for baseline techniques

In addition to the MCA-type techniques, we test on two baseline transfer learning techniques that do not make use of any MCA-type process. In the first baseline ("Alone 0-1"), we train on $\mathcal{X}_{\text{kNN}}^{\text{train}}$ and test on $\mathcal{X}_1^{\text{kNN, test}}$. In the second baseline ("Alone 0-2"), we train on $\mathcal{X}_{\text{kNN}}^{\text{train}}$ and test on $\mathcal{X}_2^{\text{kNN, test}}$.

### 4.3.3 Selected testing procedures

While SGM is capable of handling a full matched data point containing information from all three modalities simultaneously, MCA can handle only two modalities at a time. For this reason, we've compared the performance of SGM to the performance of MCA as applied to both modality pairs that include the central modality (0-1 and 0-2). We've also included in our comparisons the performance of stacked MCA, in which both non-central modalities were stacked to form a single modality, followed by an application of standard MCA. In addition, we've included the performance of the kNN as trained on the central modality, modality 0, followed by testing on each of the non-central modalities.

Below we summarize the six methods whose testing procedures are outlined in Sections 4.3.1 and 4.3.2, with the norm-connectedness of the four MCA-type techniques illustrated in Figure 4 and their training and testing procedures summarized in Figure 5.

1. Alone 0-1: No MCA-type process is applied; train kNN on modality 0; test kNN on modality 1.

2. Alone 0-2: No MCA-type process is applied; train kNN on modality 0; test kNN on modality 2.

3. MCA 0-1 (Figure 4a): Apply MCA to modalities 0 and 1; train kNN on common-domain data from modality 0; test kNN on common-domain data from modality 1.

4. MCA 0-2 (Figure 4b): Apply MCA to modalities 0 and 2; train kNN on common-domain data from modality 0; test kNN on common-domain data from modality 2.

5. Stacked MCA (Figure 4c): Apply MCA to modalities 0 and 1-2 stacked; train kNN on common-domain data from modality 0; test kNN on common-domain data from stacked modality.

6. SGM (Figure 4d): Apply SGM to modalities 0 (central), 1, and 2; train kNN on common-domain data from modality 0; test kNN on common-domain centroid data from modalities 1 and 2, with empirically chosen weights 0.2 and 0.8 for modalities 1 and 2, respectively.

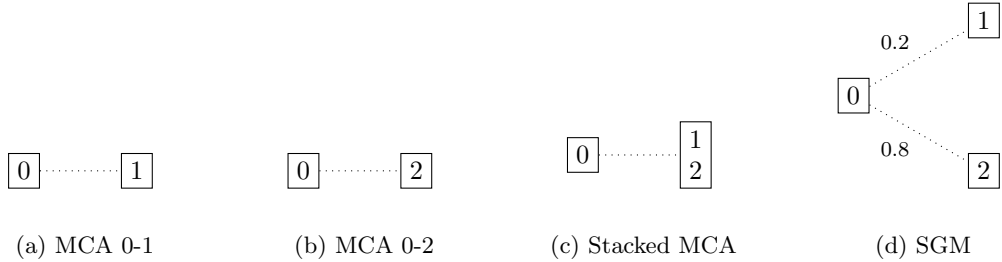

(a) MCA 0-1  (b) MCA 0-2  (c) Stacked MCA  (d) SGM

Figure 4: Schematic representation of the norm-connectedness of the four MCA-type methods compared in our numerical experiments. The modality labels here match those of Figure 2.

### 4.4 Results

In addition to kNN classifier performance, we've also plotted, as a percentage, the objective function trace ratio

$$T \equiv \frac{2\sum_{i\in[m]} \lambda_i \operatorname{tr}\left((D_i^{\text{iter}})^{\mathsf{T}} \operatorname{dg}_{r_i^{\min}}(\Sigma_{C_i})((U_{C_i})_{[r_i^{\min}]})^{\mathsf{T}}(U_{C_0})_{[r_0^{\min}]} \operatorname{dg}_{r_0^{\min}}(\Sigma_{C_0}) D_0^{\text{iter}} V_{S_0}^{\mathsf{T}} V_{S_i}\right)}{\sum_{i\in[m]} \lambda_i \operatorname{tr}\left(B_{C_0 C_0^{\mathsf{T}}}^{r_0^{\min}} + B_{C_i C_i^{\mathsf{T}}}^{r_i^{\min}}\right)} \leq 1 \qquad (33)$$

for SGM. This is the achieved fraction of the maximum possible trace, with 100% corresponding to an original objective function value (5a) of 0, indicating that all matched data points were mapped perfectly (that is, $g_0(X_0(\omega_j)) = g_i(X_i(\omega_j))$ for each $i \in [m]$ and for each $j \in [n]$). The inequality (33) comes from setting (18a) $\geq 0$ and then using (23b); it represents one measure of the quality of the solutions (27) used in the formation of the SGM maps.

Figures 6a and 6b show the performance of a kNN classifier as applied to data from each of the six methods outlined in Section 4.3.3, along with SGM's objective function trace ratio $T$ of (33). In each figure, SGM outperforms the other five methods for virtually all values of the dimension $k$ of the common domain. For small values of $n$, the non-SGM methods do not have enough training samples to produce good performance, even relative to the "Alone" baselines. We note that the value of $k$ has no impact on the "Alone" baselines, as they do not involve MCA; this is why they appear as horizontal lines.

Comparing Figures 6a and 6b, it is evident that SGM with $n = 5$ is capable of outperforming the other five methods with $n = 20$. This suggests that SGM may be a useful tool in contexts where few matched data points are available for training. Since MCA was developed with the idea that matching is an expensive process [1], a method that is performant using a low value of $n$ (that is, few MCA training samples) is desirable.

In addition, the values of $T$ in these experiments are above 80%, suggesting that the approximations used in the SGM solutions are of reasonably good quality. At this time, we do not have a rule of thumb for the minimum value of $T$ that indicates good quality. Empirically, however, the value of $T$ tends to be slightly above $\max_j\{\lambda_j\}$ when one weight dominates the others.

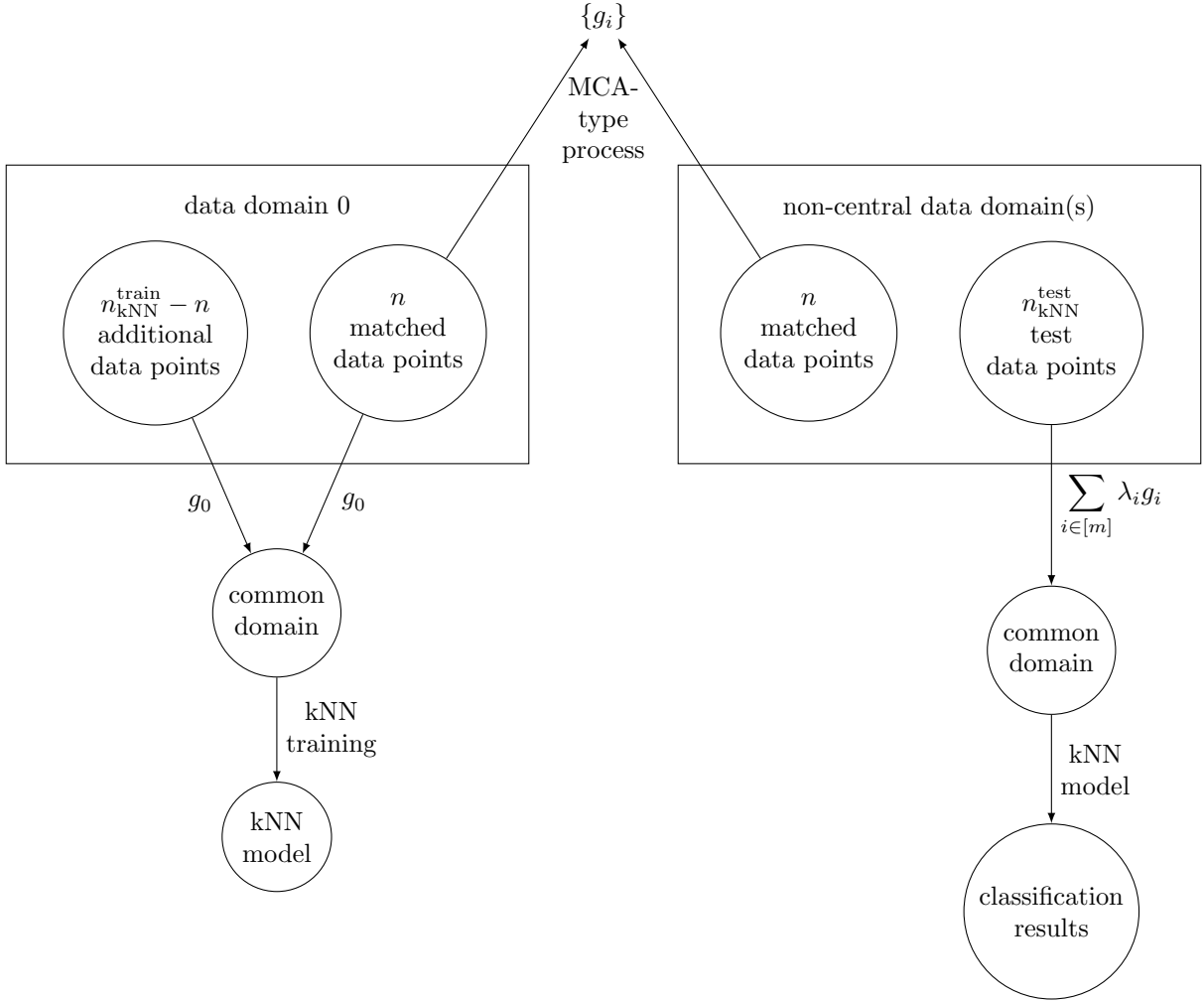

Figure 5: Schematic representation of the training and testing procedures for each MCA-type process, as outlined in Sections 4.3.1 and 4.3.3. First, an MCA-type process is applied to the matched data points, forming the maps $\{g_i\}$ (top). Then, a kNN is trained on a set of points mapped from data domain 0 to the common domain via $g_0$ (bottom left). Finally, the kNN model is applied to the common-domain centroids of a set of points mapped from the non-central data domains (bottom right). Importantly, the sets $\{\omega_j\}$ corresponding to the kNN training and testing data points are mutually disjoint.

## 5 Discussion

Our key theoretical results are the SGM maps (27) and their method of computation, as provided in Algorithm 3.1.

Our key numerical results are summarized in Figure 6, which demonstrates our claim that SGM is capable of encoding into its maps more information than MCA when few training points are available. This suggests that SGM's proper handling of multiple modalities confers some advantage in this case.

We note that the value of $r_i^{\min}$ of (17) roughly corresponds to the amount of "bandwidth" available for a given modality to influence another modality's map. In our experiments, we've enforced identity covariances in the common domain, so $\text{rank}(C_i C_i^\mathsf{T}) = k$. Since each data domain consists of points in $\mathbb{R}^{49}$ (see Section 4.1), we also have, from (13), that $\text{rank}(S_i) \leq \min\{49, n-1\}$. For the values of $n$ in the experiments of Figures 6a and 6b, then, we have that $\text{rank}(S_i) \leq n-1$, and so $r_i^{\min} \leq \{k, n-1\}$. As $k$ approaches $n-1$, then, we expect the performance of the MCA-type processes to level out. This is exactly the behavior we see in Figures 6a and 6b.

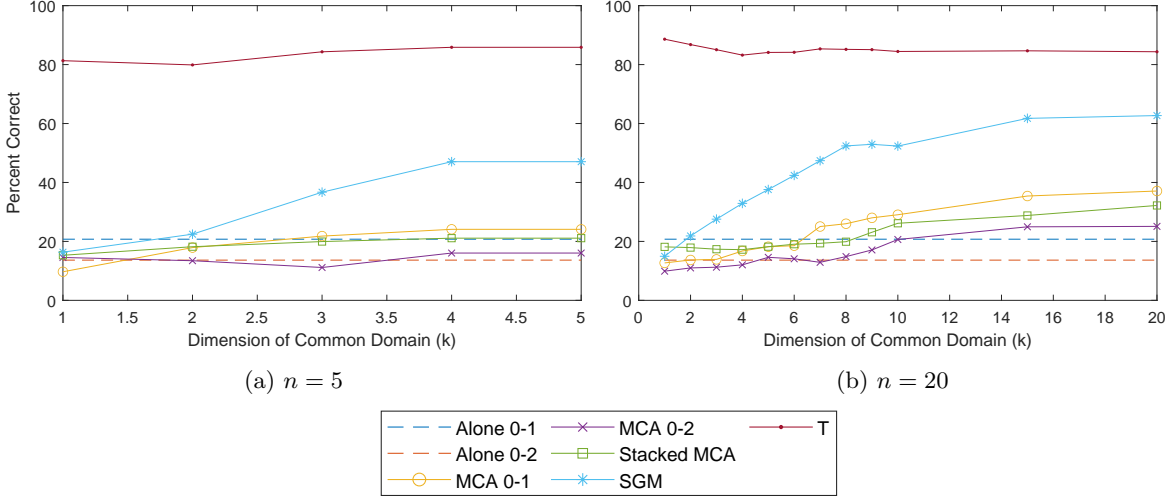

Figure 6: Performance comparison between the methods outlined in Figure 4. MCA-type maps were computed with (a) $n = 5$ and (b) $n = 20$ matched data points; percent correct is according to a kNN classifier trained with $n_{\text{kNN}}^{\text{train}} = 5,000$ points using 15 neighbors and tested with $n_{\text{kNN}}^{\text{test}} = 2,000$ points; $T$ is defined as in (33).

## 6 Conclusion

In order to extend the application space of the MCA family of techniques to the multimodal case, this paper developed SGM, a data fusion and transfer learning technique capable of directly handling multimodal data. Particular feasible points for the resulting optimization problem were found, and algorithms for their computation were provided. This generalized technique was called SGM, and was demonstrated via numerical experiments to be capable of encoding into its maps more information than MCA when few matched data points are available.

We note that the SGM technique has the potential to be useful in a variety of multimodal data fusion and transfer learning applications, particularly in data-limited contexts.

### Acknowledgments

The author would like to acknowledge support from the Air Force Research Laboratory (AFRL) through contract FA8650-19-F-5230 and from the Air Force Office of Scientific Research (AFOSR) through grant 21RXCOR037 under the Dynamic Data and Information Processing (DDIP) program.

## Appendix A  An exact solution to a related problem

We are interested in finding

$$\mathcal{Q}^{\text{opt}} \in \arg\max_{\mathcal{Q} \in \mathcal{F}_Q} \sum_{i \in [m]} \lambda_i \operatorname{tr}\left(Q_i^\mathsf{T} L_i^\mathsf{T} Q_0 R_i\right), \tag{34a}$$

where

$$\mathcal{F}_Q \equiv \{(Q_0, \ldots, Q_m) \mid Q_0^\mathsf{T} \in \mathbb{O}^{\text{rows}_L \times \text{rows}_R}, Q_i^\mathsf{T} \in \mathbb{O}^{\text{cols}(R_i) \times \text{cols}(L_i)}, i \in [m]\} \tag{34b}$$

is the set of length-$(m+1)$ tuples of feasible matrices, and where

$$\text{rows}_L \equiv \text{rows}(L_j) = \text{rows}(L_i) \leq \text{rows}(R_i) = \text{rows}(R_j) \equiv \text{rows}_R, \tag{34c}$$
$$\text{cols}(L_i) \leq \text{cols}(R_i), \tag{34d}$$
$$i, j \in [m].$$

Since finding $\mathcal{Q}^{\text{opt}}$ is a generalization of an NP-hard task [9], we instead investigate the following related problem.

We seek a particular solution $\mathcal{Q}^{\text{p}} \in \mathcal{F}_{\mathcal{Q}}$ satisfying

$$Q_0^{\text{p}} \in \arg \max_{(\hat{Q}_0, \cdot, \ldots, \cdot) \in \mathcal{F}_{\mathcal{Q}}} \sum_{i \in [m]} \lambda_i \, \text{tr}\left((Q_i^{\text{p}})^{\mathsf{T}} L_i^{\mathsf{T}} \hat{Q}_0 R_i\right), \tag{35a}$$

where

$$(Q_1^{\text{p}}, \ldots, Q_m^{\text{p}}) \in \arg \max_{(\cdot, \hat{Q}_1, \ldots, \hat{Q}_m) \in \mathcal{F}_{\mathcal{Q}}} \sum_{i \in [m]} \lambda_i \, \text{tr}\left(\hat{Q}_i^{\mathsf{T}} \bar{V}_{L_i} \bar{\Sigma}_{L_i}^{\mathsf{T}} I_{\text{rows}_L \times \text{rows}_R} \bar{\Sigma}_{R_i} \bar{V}_{R_i}^{\mathsf{T}}\right). \tag{35b}$$

Now, since $\bar{\Sigma}_{L_i}^{\mathsf{T}} I_{\text{rows}_L \times \text{rows}_R} \bar{\Sigma}_{R_i} \in \mathbb{D}_{\neq}^{\text{cols}(L_i) \times \text{cols}(R_i)}$, the matrix multiplying $\hat{Q}_i^{\mathsf{T}}$ in the trace of (35b) is already factored as an SVD; an element of the set on the RHS of (35b) is found by maximizing each term separately, recognizing each as a semi-orthogonal extension of the two-sided rotation (Procrustes) problem [6, p. 463]. Such a solution is given by [2]

$$Q_i^{\text{p}} = \bar{V}_{L_i} \left((\bar{V}_{R_i})_{[\text{cols}(L_i)]}\right)^{\mathsf{T}}, \quad i \in [m]. \tag{36}$$

With $Q_i^{\text{p}}$ chosen as in (36), we now turn our attention to (35a). Rewriting the objective function as

$$\text{tr}\left(\hat{Q}_0 H\right), \quad \text{where} \quad H \equiv \sum_{i \in [m]} \lambda_i R_i (Q_i^{\text{p}})^{\mathsf{T}} L_i^{\mathsf{T}}, \tag{37}$$

we again recognize a semi-orthogonal Procrustes problem; a solution is given by [2]

$$Q_0^{\text{p}} = \bar{V}_H \left((\bar{U}_H)_{[\text{rows}_L]}\right)^{\mathsf{T}}. \tag{38}$$

The solutions (36) and (38) are closed-form solutions achieving the upper bound [11] for the objective function of (35a) and therefore represent exact solutions to this related problem. A method to compute these solutions is provided in Algorithm A.1 below.

Algorithm A.1: Computation of an exact solution to a related problem.

---

COMPUTATION OF AN EXACT SOLUTION TO A RELATED PROBLEM

---

1 **Inputs:** matrices $\{L_i, R_i\}_{i \in [m]}$ and weights $\{\lambda_i\}_{i \in [m]}$

2 **for** $i \in [m]$

3    Compute full SVDs $L_i = \bar{U}_{L_i} \bar{\Sigma}_{L_i} \bar{V}_{L_i}^{\mathsf{T}}$ and $R_i = \bar{U}_{R_i} \bar{\Sigma}_{R_i} \bar{V}_{R_i}^{\mathsf{T}}$

4    Compute particular solution $Q_i^{\text{p}}$     See (36)

5 **end**

6 Compute matrix $H$     See (37)

7 Compute full SVD $H = \bar{U}_H \bar{\Sigma}_H \bar{V}_H^{\mathsf{T}}$

8 Compute particular solution $Q_0^{\text{p}}$     See (38)

---

To denote the result $(Q_0^{\text{p}}, \ldots, Q_m^{\text{p}})$ of applying Algorithm A.1 to the matrices $\{L_i, R_i\}_{i \in [m]}$ and weights $\{\lambda_i\}_{i \in [m]}$, we write

$$(Q_0^{\text{p}}, \ldots, Q_m^{\text{p}}) = \mathcal{A}_{A.1}(\{L_i, R_i\}_{i \in [m]}, \{\lambda_i\}_{i \in [m]}). \tag{39}$$

## Appendix B  An empirically useful iterative algorithm

We now provide an iterative algorithm that, provided a feasible starting point, sometimes increases – but never decreases – the value of the objective function of (34a). We have found this algorithm to be empirically useful.

Consider some $(Q_0, \ldots, Q_m) \in \mathcal{F}_{\mathcal{Q}}$, with $\mathcal{F}_{\mathcal{Q}}$ defined as in (34b). Let $\hat{Q}_i$ be $Q_i$ with any subset of its rows multiplied by $-1$. Then $(\hat{Q}_0, \ldots, \hat{Q}_m) \in \mathcal{F}_{\mathcal{Q}}$. In addition, observe that the objective function of (34a) may be written as

$$\sum_{i \in [m]} \lambda_i \sum_{j \in [\text{rows}(Q_i)]} \text{row}_j(L_i^\mathsf{T} Q_0 R_i) \cdot \text{row}_j(Q_i). \tag{40}$$

Provided with feasible $(Q_0, \ldots, Q_m)$, then, any time the dot product in (40) is negative, the value of the objective function (40) can be improved by multiplying $\text{row}_j(Q_i)$ by $-1$. This can be repeated for each dot product. Once this is done, we may re-compute $H$ of (37) and obtain an update for $Q_0$. This process can then be repeated; the entire process is formalized in Algorithm B.1 below.

Algorithm B.1: An empirically useful iterative algorithm.

---
AN EMPIRICALLY USEFUL ITERATIVE ALGORITHM
---

1  **Inputs:** feasible point $(Q_0, \ldots, Q_m) \in \mathcal{F}_{\mathcal{Q}}$, matrices $\{L_i, R_i\}_{i \in [m]}$, and weights $\{\lambda_i\}_{i \in [m]}$
2  count $\leftarrow 1$
3  **while** count $> 0$
4      count $\leftarrow 0$
5      **for** $i \in [m]$
6          **for** $j \in [\text{rows}(Q_i)]$
7              **if** $\text{row}_j(L_i^\mathsf{T} Q_0 R_i) \cdot \text{row}_j(Q_i) < 0$
8                  $\text{row}_j(Q_i) \leftarrow -\text{row}_j(Q_i)$
9                  count $\leftarrow$ count $+ 1$
10             **end**
11         **end**
12     **end**
13     **if** count $> 0$
14         Compute matrix $H = \sum_{i \in [m]} \lambda_i R_i Q_i^\mathsf{T} L_i^\mathsf{T}$
15         Compute full SVD $H = \bar{U}_H \bar{\Sigma}_H \bar{V}_H^\mathsf{T}$
16         $Q_0 \leftarrow \bar{V}_H((\bar{U}_H)_{[\text{rows}_L]})^\mathsf{T}$
17     **end**
18 **end**

---

To denote the result $(Q_0^{\text{iter}}, \ldots, Q_m^{\text{iter}})$ of applying Algorithm B.1 to the initial feasible point $(Q_0, \ldots, Q_m)$, matrices $\{L_i, R_i\}_{i \in [m]}$, and weights $\{\lambda_i\}_{i \in [m]}$, we write

$$(Q_0^{\text{iter}}, \ldots, Q_m^{\text{iter}}) = \mathcal{A}_{B.1}((Q_0, \ldots, Q_m), \{L_i, R_i\}_{i \in [m]}, \{\lambda_i\}_{i \in [m]}). \tag{41}$$